\documentclass{article}

\usepackage[numbers,compress]{natbib}
\usepackage[utf8]{inputenc}
\usepackage[T1]{fontenc}
\usepackage{mathptmx}
\usepackage[margin=1in]{geometry}
\usepackage[hypertexnames=false]{hyperref}
\usepackage{url}
\usepackage{booktabs}
\usepackage{amsfonts}
\usepackage{nicefrac}
\usepackage{microtype}
\usepackage{xcolor}
\usepackage{graphicx}
\usepackage{float}
\usepackage{amsmath}
\usepackage{algorithm}
\usepackage{algorithmic}
\usepackage{multirow}

\newcommand{\TableStd}{\renewcommand{\arraystretch}{1.08}\setlength{\tabcolsep}{5pt}}
\newcommand{\TableDense}{\renewcommand{\arraystretch}{1.08}\setlength{\tabcolsep}{4pt}}
\newcommand{\TableLoose}{\renewcommand{\arraystretch}{1.08}\setlength{\tabcolsep}{6pt}}

\title{Claim-Selective Certification for High-Risk Medical Retrieval-Augmented Generation}

\author{
  Shao Kan \\
  Jinglue Technology Development (Nanjing) Co., Ltd. \\
  Room 1215-13, 12th Floor, Building A2, Huizhi Science and Technology Park,\\
  No. 8 Hengtai Road, Nanjing Economic and Technological Development Zone,\\
  Nanjing, China \\
  \texttt{shaokan1991@gmail.com} \\
  \url{https://orcid.org/0009-0003-4872-6193}
}

\begin{document}

\maketitle

\begin{abstract}
Medical RAG systems in high-risk QA settings are often evaluated through a single answer-or-abstain decision, but mixed evidence may support one claim, require conditions for another, and contradict a third. We study \textit{claim-selective certification}: each response is decomposed into verifiable claims, scored against retrieved evidence, and mapped by an intent-aware selector to \{full, partial, conflict, abstain\}.
On the primary weak-label certificate protocol, whose real-source-only dev/test rows cover the naturally occurring non-abstain actions, the full system records UCCR$=0.0000$, PAU$=1.0000$, PAU Precision$=0.9901$, and action accuracy$=0.9204$ on dev ($n=314$), and UCCR$=0.0000$, PAU$=0.9967$, PAU Precision$=0.9739$, and action accuracy$=0.8997$ on test ($n=319$).
UCCR measures unsupported-claim risk within the certificate definition, and a source-missing counterfactual slice evaluates \texttt{abstain} under empty evidence. Shortcut controls quantify the action-label prior explained by source and intent metadata, while source/evidence-novel slices characterize transfer boundaries. The resulting interface separates action-label prediction from evidence-linked claim selection under mixed evidence.
\end{abstract}

\section{Introduction}

Retrieval-augmented generation (RAG) grounds language models in external knowledge~\cite{lewis2020rag,gao2023rag}, but high-risk medical QA has an asymmetric answer contract: unsupported safety, dosing, or contraindication claims can be harmful, while blanket abstention can suppress usable evidence. The same evidence set may support a dosing constraint, leave monitoring uncertain, and contradict a contraindication. We study \textit{claim-selective certification}: decompose a question into verifiable claim units, score each claim against evidence, and choose whether to state it directly, state it with conditions, contest it, or withhold it.

The implementation has three stages: template-based claim decomposition, cue-based relation scoring for support, conflict, and limitation signals, and an \textit{intent-aware risk-calibrated selector}. The selector receives a \texttt{question\_intent}: pregnancy, lactation, monitoring, and special-population questions often require condition-limited wording, whereas contraindication and interaction questions more often require certification or conflict handling.

We evaluate this interface on 2,223 medical QA samples, including 2,103 examples (94.6\%) from publicly downloaded real sources and 120 synthetic examples (5.4\%). Primary results use a real-source-only split, with synthetic, source-missing, and source/evidence-novel slices characterizing behavior under synthetic examples, empty evidence, and source/evidence novelty. We compare threshold-only, binary-form, NLI-based, learned-relation, and full claim-selective systems under the same weak-label certificate protocol. The study contributes a claim-level certification/action formulation, an intent-aware fixed policy layer, and a certificate-producing evaluation framework that reports UCCR, PAU, PAU Precision, action accuracy, risk--coverage, source-overlap, shortcut, and source/evidence-novel analyses for high-risk medical RAG.

\section{Related Work}
\label{sec:related}

Retrieval-augmented generation grounds language models by retrieving external documents and conditioning generation on them~\cite{lewis2020rag,izacard2021fid,gao2023rag}. Much of the recent work improves evidence acquisition through dense retrieval~\cite{karpukhin2020dpr}, iterative retrieval and self-reflection~\cite{asai2023selfrag}, query rewriting for RAG pipelines~\cite{ma2023query}, or decomposition strategies for compositional questions~\cite{press2023measuring}. RAG methods determine which evidence is available to the model. Our setting starts from the next decision point: once evidence is available, which claims should appear in the answer?

Selective prediction and abstention reduce risk by withholding low-confidence predictions~\cite{geifman2017selective,kamath2020selective}. Related QA and dialogue work studies time-sensitive answerability~\cite{chen2021time_sensitive_qa}, calibrated belief-state distributions~\cite{vanniekerk2020knowing}, and uncertainty over meanings rather than surface strings~\cite{kuhn2023semantic}. These approaches are effective when an instance is globally answerable, unanswerable, or uncertain. They are less expressive when a medical question contains several claims with different evidence states. We keep the risk-control objective but move the decision unit from the whole response to individual claims and action types.

Faithfulness and fact-checking methods verify whether generated or proposed claims are supported by sources~\cite{maynez2020faithfulness,dziri2022faithdial,rashkin2023attribution,thorne2018fever,schuster2021get}, while citation-grounded generation work studies how models present attributed evidence in the output itself~\cite{gao2023enabling}. More directly, long-form factuality evaluators decompose responses into atomic facts and check evidence support~\cite{min2023factscore,wei2024longformfactuality}, while efficient grounded checkers such as MiniCheck train smaller models for document-grounded fact verification~\cite{tang2024minicheck}. These works make factual support measurable at the claim or fact level. Claim-selective RAG uses the same kind of signal earlier in the pipeline, before the final response is formed, and maps each claim to an explicit action and certificate state rather than only scoring a completed answer.

Diagnostic evaluation work shows that high aggregate accuracy can hide shortcut behavior, annotation artifacts, and distribution-shift failures. Prior studies expose such behavior with hypothesis-only artifact analyses~\cite{gururangan2018annotation}, controlled heuristic challenge sets~\cite{mccoy2019right}, behavioral test suites~\cite{ribeiro2020checklist}, data maps~\cite{swayamdipta2020dataset}, and in-the-wild distribution-shift benchmarks~\cite{koh2021wilds}. We bring the same perspective to claim-selective RAG: shortcut controls, source/evidence-novel slices, and source-missing abstention evaluations separate certificate-producing behavior from label-space priors and transfer failures.

Medical QA has been studied in retrieval, benchmark, and large-model settings~\cite{abacha2019overview,jin2019pubmedqa,jin2021disease,pal2022medmcqa,singhal2023clinicalknowledge,nori2023capabilities}. Many datasets evaluate multiple-choice, yes/no, or complete-answer behavior. Practical medical questions, however, often need partial support, contraindication handling, or condition-specific wording. The present work targets that interface gap by evaluating claim-level selection, explicit action calibration, and risk--utility metrics such as UCCR and PAU.

\section{Problem Formulation}
\label{sec:problem}

Let $q$ denote a medical question and let $\mathcal{E}=\{e_1,\ldots,e_k\}$ be the retrieved evidence available to the system. A conventional RAG policy emits one response $r=\mathrm{LM}(q,\mathcal{E})$ and is then evaluated as a whole. This is too coarse for mixed evidence: the same evidence packet can support one claim, only conditionally support a second, contradict a third, and leave a fourth unsupported. We instead treat the response as a decision over claims.

The interface first produces a small set of verifiable claim skeletons $\mathcal{C}=\{c_1,\ldots,c_n\}$ and metadata $m_i$ for each claim, including question intent and claim type. Each claim--evidence pair receives a relation vector
\begin{equation}
\mathbf{s}(c_i,e_j)=
\big(s_{\mathrm{sup}},s_{\mathrm{conf}},s_{\mathrm{lim}}\big)\in[0,1]^3 ,
\end{equation}
where the coordinates summarize support, conflict, and limitation/conditionality. Scores are pooled across evidence with max aggregation:
\begin{equation}
S_\ell(c_i)=\max_{e_j\in\mathcal{E}} s_\ell(c_i,e_j),
\qquad
\ell\in\{\mathrm{sup},\mathrm{conf},\mathrm{lim}\}.
\end{equation}
The selector maps $(S_{\mathrm{sup}},S_{\mathrm{conf}},S_{\mathrm{lim}},m_i)$ to a claim status
\begin{equation}
z_i\in\{\mathrm{certified},\mathrm{condition\mbox{-}limited},\mathrm{conflicting},\mathrm{omitted}\}.
\end{equation}
Certified and condition-limited claims may be expressed with evidence links; conflicting claims trigger conflict behavior; omitted claims are withheld.

The final action is a coarse summary of the selected claim set:
\begin{equation}
a\in\{\texttt{full},\texttt{partial},\texttt{conflict},\texttt{abstain}\}.
\end{equation}
Intuitively, \texttt{full} means the requested answer can be stated directly, \texttt{partial} means usable information exists but requires caveats or omissions, \texttt{conflict} means material evidence disagreement or risk should be surfaced, and \texttt{abstain} means no claim should be expressed. The action is therefore not a free-form answer label; it is a policy decision induced by claim-level statuses and evidence certificates.

Evaluation follows the same claim interface. Let $\mathcal{P}$ be the expressed predicted claims and $\mathcal{G}_{\mathrm{use}}$ the construction-derived gold-usable claims under the weak-label protocol. Unsupported Critical Claim Rate (UCCR) measures certificate-level risk among expressed critical claims:
\begin{equation}
\mathrm{UCCR}=
\frac{|\{c\in\mathcal{P}: c\text{ is critical and }c\notin\mathcal{G}_{\mathrm{use}}\}|}
{|\{c\in\mathcal{P}: c\text{ is critical}\}|},
\end{equation}
defined as zero when no critical claim is expressed. The target is zero unsupported expressed critical claims. Partial Answer Utility (PAU) measures retained usable information:
\begin{equation}
\mathrm{PAU}=
\frac{|\mathcal{P}\cap\mathcal{G}_{\mathrm{use}}|}
{|\mathcal{G}_{\mathrm{use}}|}.
\end{equation}
We also report PAU Precision and F1 to distinguish retaining usable claims from over-expressing extra claims, and Action Accuracy to evaluate the four-way action interface. These are protocol metrics over weak labels, not expert clinical judgments.

This formulation makes the central tradeoff explicit. If the evidence supports $\mathcal{C}_{\mathrm{sup}}\subset\mathcal{C}$ and does not support the remaining claims, a document-level answer policy largely chooses between expressing everything, which can raise UCCR, or abstaining, which lowers PAU. A claim-selective policy can instead express $\mathcal{C}_{\mathrm{sup}}$, omit unsupported claims, and return \texttt{partial} when appropriate. The experiments ask whether this interface can meet the certificate target while preserving utility and whether the resulting actions remain meaningful beyond shortcut priors.

\section{Method}
\label{sec:method}

\begin{figure}[t]
\centering
\includegraphics[width=0.78\linewidth]{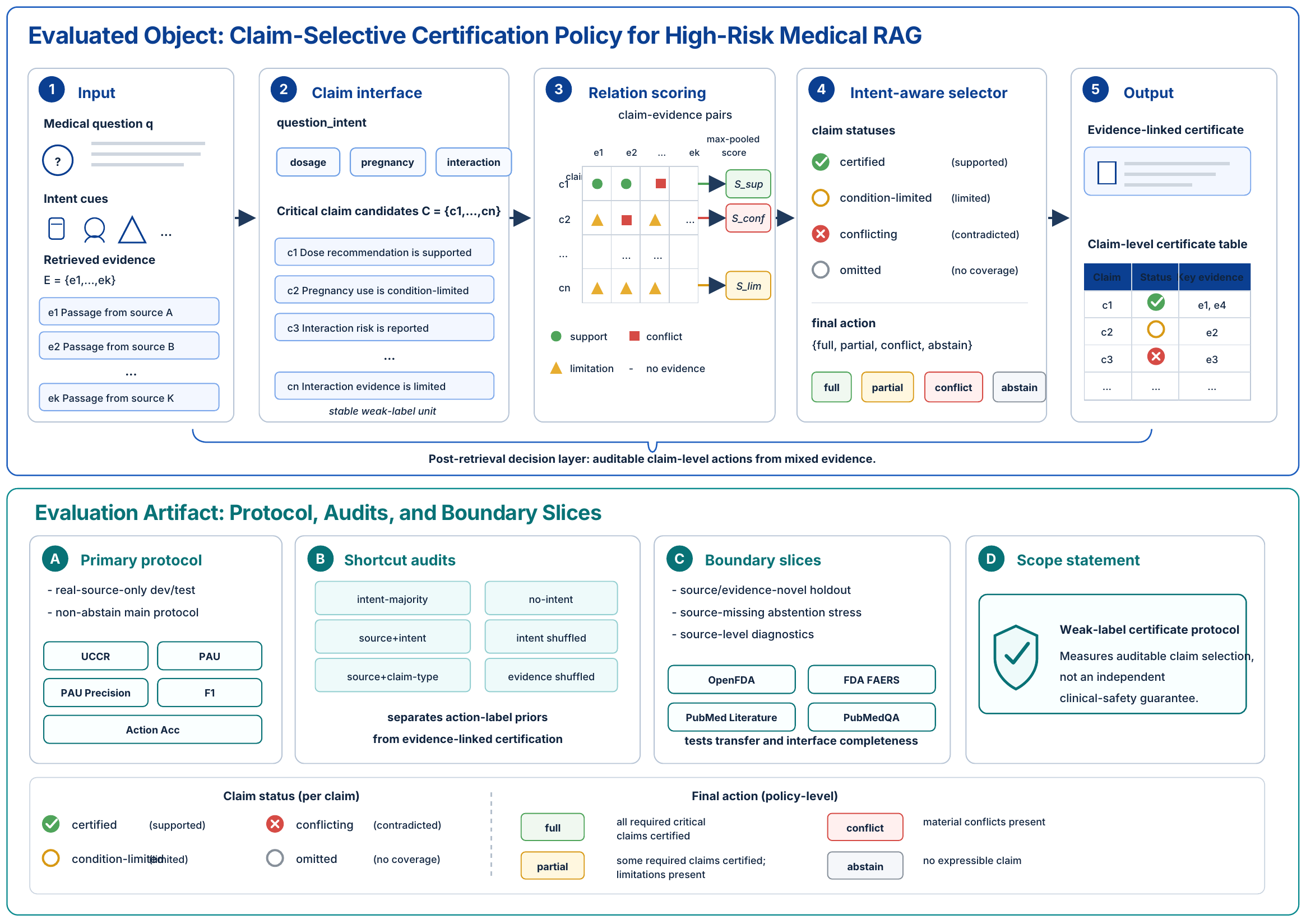}
\caption{System architecture. The pipeline combines template-based claim decomposition with explicit question intent, cue-based relation scoring, and an intent-aware risk-calibrated selector.}
\label{fig:architecture}
\end{figure}

The system follows a three-stage pipeline: claim decomposition, relation scoring, and intent-aware risk-calibrated selection. We study the post-retrieval decision layer: given a fixed evidence set, the system decides which claims should be certified, condition-limited, contested, or omitted. Figure~\ref{fig:architecture} summarizes the pipeline. The system exposes a small set of candidate claims, scores each claim against retrieved evidence, and applies an explicit action policy. This keeps the main decisions inspectable under weak supervision and makes it possible to evaluate action calibration separately from open-ended generation.

\subsection{Design Rationale}
\label{sec:design_rationale}

The design centers the post-retrieval decision boundary: the claim interface fixes the decision unit, the relation scorer exposes support, conflict, and limitation signals, and the selector maps those signals to an action. This structure lets us study action calibration under a common weak-label protocol while keeping claim selection, certification, and audit tied to explicit intermediate variables.

\subsection{Claim Decomposition}
\label{sec:decomposition}

Given a query $q$, the decomposition layer emits verifiable claim skeletons $\mathcal{C} = \{c_1, \ldots, c_n\}$. The templates produce a small set of high-level critical claims rather than unrestricted fine-grained extractions. Each query also receives a \texttt{question\_intent}, covering indication, dosage, contraindication, interaction, pregnancy/lactation, monitoring, missed-dose, and special-population questions. The resulting skeletons provide a stable interface for studying selective inclusion, condition-limited wording, conflict handling, and omission under a shared weak-label protocol.

\begin{table}[t]
\centering
\caption{Claim-template and selector-policy interface. The rows summarize the operating families used by the selector.}
\label{tab:method_policy_families}
\small
\TableDense
\begin{tabular}{p{0.33\linewidth}p{0.59\linewidth}}
\toprule
\textbf{Question group} & \textbf{Claim skeleton and policy family} \\
\midrule
Indication / effectiveness & Drug is indicated or supported for a condition; certify only with support, otherwise mark as condition-limited or omit. \\
Pregnancy, lactation, special population & Use is appropriate under a population-specific condition; prefer condition-limited wording when limitation cues are present. \\
Contraindication / interaction & Evidence reports a prohibition, interaction, or material risk; disclose conflict/risk rather than directly certify. \\
Dosage, missed dose, monitoring & A specific dosing or monitoring action is supported; require explicit instruction-level support. \\
\bottomrule
\end{tabular}
\end{table}

\subsection{Relation Scoring}
\label{sec:relation}

For each claim--evidence pair $(c_i, e_j)$, the relation module estimates support, conflict, and limitation scores. The scorer combines lexical overlap with a cue lexicon to identify whether evidence supports a claim, contradicts it, or supports it only with conditions. The cue lexicon contains English phrases from labels and abstracts. Scores are aggregated across retrieved evidence with max-style pooling so that one strong passage can activate the relevant relation signal:
\begin{align}
S_{\text{support}}(c_i) &= \max_{e_j \in \mathcal{E}} s_{\text{support}}(c_i, e_j) \\
S_{\text{conflict}}(c_i) &= \max_{e_j \in \mathcal{E}} s_{\text{conflict}}(c_i, e_j) \\
S_{\text{limitation}}(c_i) &= \max_{e_j \in \mathcal{E}} s_{\text{limitation}}(c_i, e_j).
\end{align}
The scorer preserves the originating \texttt{question\_intent}, so the selector can condition on both evidence relations and question type.

\subsection{Intent-Aware Risk-Calibrated Selection}
\label{sec:selection}

The selector maps each scored claim to one of four statuses: \textbf{certified}, when evidence is strong enough for direct inclusion; \textbf{condition-limited}, when the claim should be used only with caveats or partial wording; \textbf{conflicting}, when the evidence is contradictory or too risky for direct certification; and \textbf{omitted}, when support is too weak. The final response action is chosen from \{full, partial, conflict, abstain\} based on the selected claim set.

The mapping is intent-aware. A single global threshold is not adequate across medical question types: monitoring, pregnancy, lactation, and special-population questions often require condition-limited answers even when some support is present, whereas contraindication and interaction questions more often require certification or conflict handling. The selector groups intents into coarse policy families---full-certify oriented, partial-oriented, conflict-oriented, mixed, and dosage-specific cases---and applies different mappings from $(S_{\text{support}}, S_{\text{conflict}}, S_{\text{limitation}})$ to claim status. The ablation compares this policy with a threshold-only selector that uses generic score thresholds without intent-specific rules.

The implementation also contains two scoped source-family priors. First, explicit dosage-instruction cues from openFDA label evidence can rescue dosage support in the relation scorer. Second, PubMed Literature review-level evidence can downgrade selected full answers to partial for specified claim families. These priors are fixed implementation choices, not learned calibration parameters; the experiments therefore report source-conditioned majority controls and source/evidence-novel slices to expose how much action behavior is recoverable from source and intent metadata.

\section{Experiments}
\label{sec:experiments}

We evaluate certificate risk, retained utility, intent-aware action calibration, expressiveness beyond answer/abstain baselines, and behavior under source/evidence novelty.

\subsection{Data Provenance and Main Evaluation Split}
\label{sec:dataset}

We use a 2,223-item medical QA collection: 2,103 public-source examples (94.6\%) and 120 synthetic examples (5.4\%). Main results use a real-source-only split with train $n=1{,}470$, dev eval $n=314$, and test eval $n=319$; the synthetic subset contributes a separate $n=20$ stress slice. We also derive a source/evidence-novel holdout from primary dev/test rows whose normalized \texttt{source\_url} and \texttt{evidence\_text} are both absent from train (dev $n=82$, test $n=78$). This slice evaluates source/evidence novelty, while the primary split remains the main protocol. A source-missing counterfactual slice preserves questions and claim skeletons, removes evidence, and sets gold action to \texttt{abstain}, targeting deterministic retrieval-failure abstention under empty evidence. Tables~\ref{tab:appendix_clean_mainline}, \ref{tab:appendix_abstain_scope}, and~\ref{tab:source_novel_holdout} summarize counts.

PubMedQA maps abstract-level \texttt{yes}/\texttt{no}/\texttt{maybe} research conclusions into \texttt{full}/\texttt{conflict}/\texttt{partial}, so it serves as an abstract-style interface-transfer slice rather than a drug-label QA proxy.

\subsection{Metrics}
\label{sec:metrics}

We report four metrics. \textbf{UCCR} measures the fraction of expressed critical claims that are unsupported under the certificate definition. \textbf{PAU} is recall-style utility over retained gold-usable claims and is paired with PAU Precision, F1, and risk--coverage curves to expose over-expression. \textbf{F1} uses sample-scoped claim matching because claim identifiers are reused. \textbf{Action Accuracy} evaluates \{full, partial, conflict, abstain\} actions and is interpreted together with certificate production and perturbation controls.

\subsection{Weak-Label Certificate Protocol}
\label{sec:weak_label_scope}

Weak labels define a shared certificate protocol: each prediction records claim status, evidence identifiers, and relation scores. UCCR and PAU are protocol measurements over expressed unsupported critical claims and retained usable claims. Shortcut controls and risk--coverage curves are reported alongside them so that action behavior, certificate production, and label-space priors are analyzed together. To calibrate this protocol against external review, we separately re-audited a 100-item human-validation subset using official-source-first evidence review with preserved screenshots, HTML snapshots, and text traces. The final audit labels contain 49 \texttt{full\_support}, 49 \texttt{conditional\_support}, and 2 \texttt{conflict} cases, agreeing with the weak labels on 73/100 items (0.7300; Cohen's $\kappa=0.5027$). This subset is not used for threshold tuning or benchmark replacement; it is reported only to bound how closely the weak-label certificate protocol tracks pharmacist adjudication.

\subsection{Evaluation Protocol and Label--Policy Separation}
\label{sec:label_policy_separation}

All numbers are computed against construction-derived weak labels. We report dev and test results on the primary real-source-only split; appendix material provides split details, commands, intervals, and risk--coverage analysis. The threshold-only baseline is tuned on \texttt{dev\_eval} by a small grid search over \texttt{support}, \texttt{conflict}, and \texttt{condition\_limited}; among UCCR$=0$ candidates, we maximize PAU, then Action Accuracy, then F1. The selected \texttt{support}=0.35, \texttt{conflict}=0.55, and \texttt{condition\_limited}=0.30 thresholds transfer unchanged to test. The full selector is a fixed policy specification with intent-conditioned branches and global fallback thresholds, analyzed with shortcut controls, threshold perturbations, and a policy-constant audit. The speech-act-guided proxy tunes one global \texttt{answer\_support}=0.34 gate on dev and transfers it unchanged to test.

At inference, the full selector predicts from the question, claim skeleton and intent, retrieved evidence, relation scores, and documented source/context fields used by scoped source-family priors. Because source type, source-level claim type, and intent can carry action priors, we report metadata-only majority controls fit on train; Table~\ref{tab:appendix_label_policy_info} lists label-policy information access.

The analysis covers retrieval-only, threshold-only, shortcut, binary-form, NLI, learned-relation, and full claim-selective rows. Binary-form baselines collapse the final action space to answer/abstain; NLI and learned-relation rows keep the native \{full, partial, conflict, abstain\} interface, with the learned row swapping only the relation module.

\section{Results}
\label{sec:results}

\subsection{Main Ablation Results}

We first report primary real-source-only results, then use controls and transfer slices to separate certificate production from action-label priors. Table~\ref{tab:main_ablation} isolates the transition from evidence access to relation scoring and then to action-level policy.

\begin{table*}[t]
\centering
\caption{Main ablation results on the primary real-source-only split. UCCR is defined by the weak-label certificate protocol (Section~\ref{sec:weak_label_scope}), and PAU Prec is the fraction of expressed claims that are gold-usable.}
\label{tab:main_ablation}
\small
\TableLoose
\begin{tabular}{llccccc}
\toprule
\textbf{Split} & \textbf{Configuration} & \textbf{UCCR} & \textbf{PAU} & \textbf{PAU Prec} & \textbf{F1} & \textbf{Action Acc} \\
\midrule
\multirow{4}{*}{Dev ($n=314$)}
& Retrieval only & 0.0860 & 1.0000 & 0.9522 & 0.9755 & 0.4968 \\
& Relation only & 0.1943 & 1.0000 & 0.9522 & 0.9755 & 0.4968 \\
& Threshold-only selector & 0.0000 & 0.9933 & 0.9581 & 0.9754 & 0.5223 \\
& Full risk-calibrated & \textbf{0.0000} & \textbf{1.0000} & \textbf{0.9901} & \textbf{0.9950} & \textbf{0.9204} \\
\midrule
\multirow{4}{*}{Test ($n=319$)}
& Retrieval only & 0.0878 & 1.0000 & 0.9373 & 0.9676 & 0.5862 \\
& Relation only & 0.2069 & 1.0000 & 0.9373 & 0.9676 & 0.5862 \\
& Threshold-only selector & 0.0000 & 0.9732 & 0.9510 & 0.9620 & 0.5517 \\
& Full risk-calibrated & \textbf{0.0000} & \textbf{0.9967} & \textbf{0.9739} & \textbf{0.9851} & \textbf{0.8997} \\
\bottomrule
\end{tabular}
\end{table*}

Retrieval-only and relation-only rows express available claims without action selection, producing nonzero UCCR (0.0860/0.0878 and 0.1943/0.2069 on dev/test) and low action accuracy. The dev-tuned threshold-only selector restores UCCR to zero and reaches PAU 0.9933/0.9732 and F1 0.9754/0.9620, but action accuracy remains 0.5223/0.5517. The full selector keeps UCCR$=0.0000$ while improving PAU, PAU Precision, F1, and action accuracy to 0.9204/0.8997. Bootstrap intervals keep the full selector's F1 and Action Accuracy gains over threshold-only positive on both splits; risk--coverage analysis gives the same within-system pattern. The separate 100-item human re-audit shows that the weak-label protocol is informative but not expert-equivalent: weak and audited labels agree on 73/100 items with $\kappa=0.5027$, and the disagreement mass is concentrated in \texttt{full}/\texttt{conditional} boundary cases rather than support-versus-conflict reversals. Table~\ref{tab:main_ablation} should therefore be read as a protocol-level certificate and action comparison, not as a substitute for expert clinical adjudication.

\begin{figure}[t]
\centering
\includegraphics[width=\textwidth]{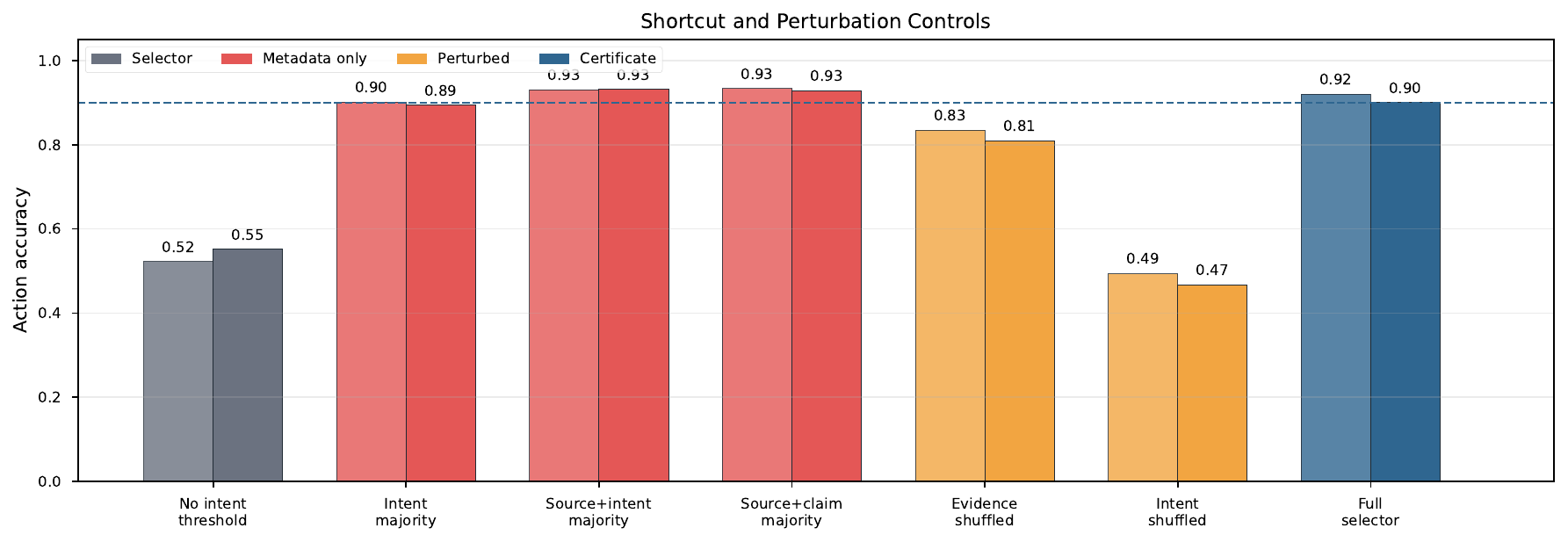}
\caption{Shortcut and perturbation controls on the primary split. Metadata-only majority rows are action-only controls fit from training-set weak labels, and certificate metrics apply only to evidence-linked claim outputs.}
\label{fig:shortcut_controls_main}
\end{figure}

Shortcut controls reveal a strong action-label prior. Intent-majority reaches 0.9013/0.8934 action accuracy without evidence; source+intent and source+claim-type action-only rows reach 0.9299/0.9310 and 0.9331/0.9279, exceeding the full selector on action accuracy alone. These rows do not select claims, assign evidence, expose relation scores, or produce certificates, so UCCR and PAU are not applicable. Perturbations confirm that the full selector still uses policy inputs: removing intent gives 0.5223/0.5517, shuffling intent gives 0.4936/0.4671, and shuffling evidence gives 0.8344/0.8088.

\subsection{External-Form Baselines}

The ablation table tests the within-system contribution chain. We add response-format baselines: direct answering, binary answer/abstain behavior, a speech-act-guided answer/abstain proxy, a standard NLI-based semantic claim-selective baseline, and a stronger learned-relation claim-selective variant. Full numeric rows are reported in Table~\ref{tab:appendix_baseline_package}.

\begin{figure}[t]
\centering
\includegraphics[width=\textwidth]{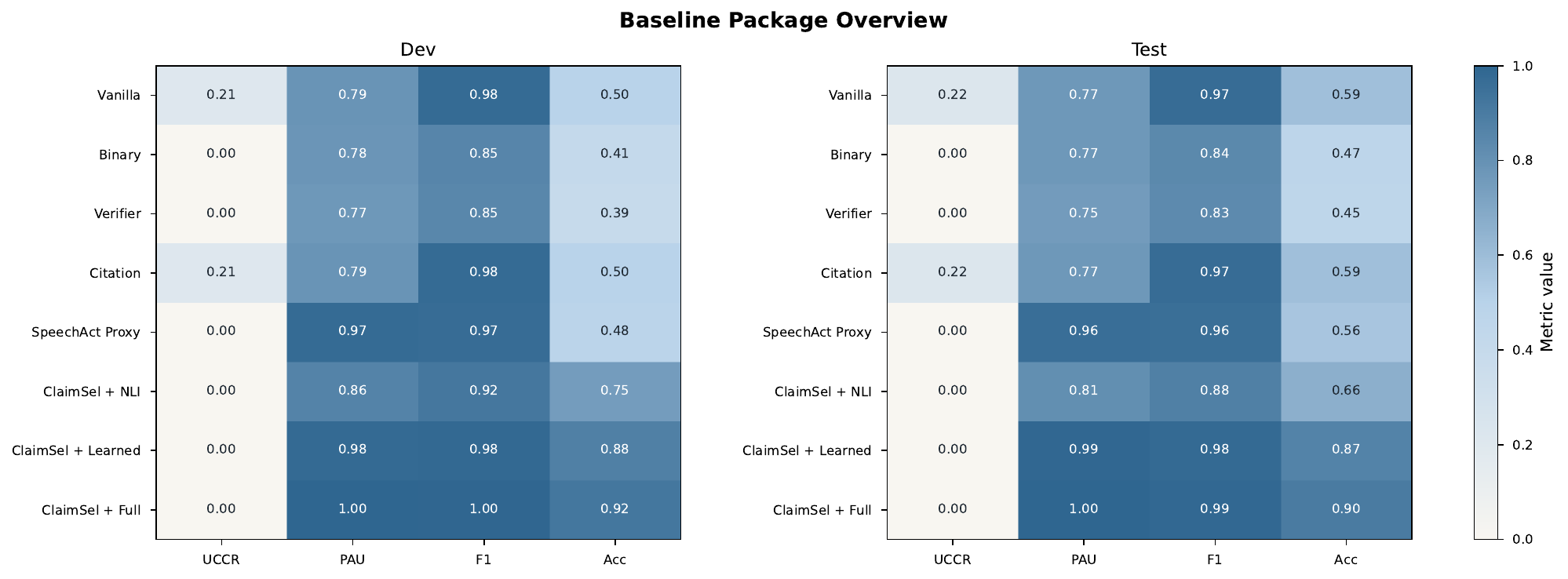}
\caption{Baseline operating map on the primary split. Binary-form baselines reduce unsupported-claim risk by collapsing the action space, whereas claim-selective baselines retain the native action interface.}
\label{fig:baseline_overview_main}
\end{figure}

Direct and citation-only answers incur large UCCR penalties. Binary answer/abstain and verifier-only rows drive UCCR to zero but sacrifice PAU and action accuracy by collapsing \texttt{partial} and \texttt{conflict}. The speech-act-guided proxy is stronger, reaching PAU 0.9732/0.9599 and F1 0.9652/0.9551 with UCCR zero, but action accuracy remains 0.4777/0.5611. The NLI-based claim-selective baseline improves action accuracy to 0.7484/0.6614. The strongest external baseline swaps the relation module while retaining the same intent-aware selector and native claim-selective action space; it reaches 0.8758/0.8652 action accuracy. A controlled relation-module comparison shows action-accuracy changes of $-0.0446$ on dev and $-0.0345$ on test, placing the main contribution at the claim-selective action policy.

\subsection{Abstract-Style Transfer Behavior and Failure Modes}

Source slices localize the remaining errors (Table~\ref{tab:appendix_source_slices}). OpenFDA and FDA FAERS are stable; PubMed Literature is mostly handled by the partial-answer policy. PubMedQA is the main abstract-style transfer slice: its \texttt{yes}/\texttt{no}/\texttt{maybe} research judgments do not map cleanly onto \texttt{full}/\texttt{conflict}/\texttt{partial}, and test action accuracy is 0.5161.

\begin{figure}[t]
\centering
\includegraphics[width=0.78\textwidth]{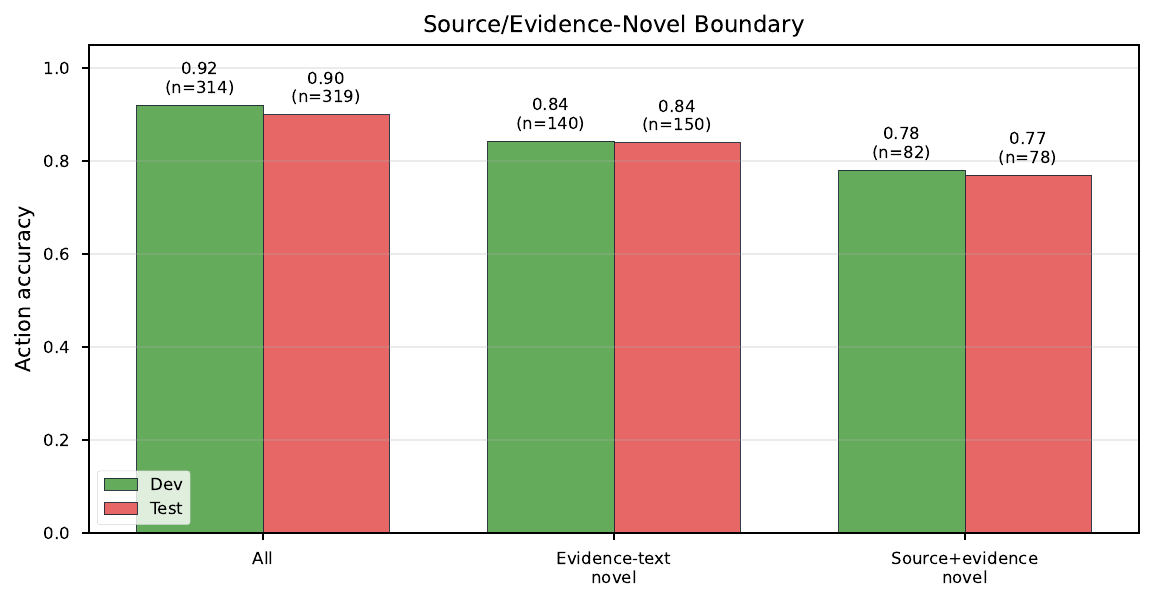}
\caption{Source/evidence-novel boundary. The full selector keeps the certificate target as overlap constraints tighten, but action accuracy drops on the strict source/evidence-novel slice.}
\label{fig:source_novel_boundary}
\end{figure}

The primary split has no exact question overlap but substantial source reuse: train-to-dev/test source-URL overlap is 0.7389/0.7555 and evidence-text overlap is 0.5541/0.5298. We therefore materialize the strictest still-usable source/evidence-novel holdout. The full selector keeps UCCR$=0.0000$ and PAU$=1.0000$, but action accuracy drops from 0.9204/0.8997 to 0.7805 on dev ($n=82$) and 0.7692 on test ($n=78$). It remains above unchanged threshold-only (0.5854/0.5256) and learned-relation swap (0.6341/0.5769), while source-conditioned majority rows remain competitive or stronger on action accuracy alone. Excluding PubMedQA, source/evidence-novel action accuracy is 0.9245 ($49/53$) on dev and 0.9362 ($44/47$) on test; the aggregate drop is concentrated in PubMedQA (0.5172/0.5161). A source-missing counterfactual slice evaluates the \texttt{abstain} action under empty evidence. Both threshold-only and full selectors abstain for all dev/test examples ($n=314/319$), giving action accuracy 1.0000; retrieval-only has action accuracy 0.0000 and UCCR$=1.0000$. The matching threshold-only result places this slice as action-space coverage evidence for the fourth action. Claim-type and error slices in the appendix show the same pattern: remaining errors are mostly \texttt{full}/\texttt{partial}/\texttt{conflict} boundary decisions under ambiguous evidence.

\section{Discussion}
\label{sec:discussion}

The results support the claim-selective action interface. Under a shared certificate protocol, it calibrates primary \texttt{full}, \texttt{partial}, and \texttt{conflict} behavior while the source-missing counterfactual slice exercises \texttt{abstain} under empty evidence. A dev-tuned threshold-only selector reaches the same UCCR target and abstains under empty evidence, but remains much weaker on primary non-abstain action accuracy.

For high-risk medical QA, the design implication is that answer selection and claim selection should be evaluated separately. A system may have enough evidence to state a dosing constraint, insufficient evidence for monitoring advice, and contradictory evidence for a contraindication in the same retrieved packet. A document-level answer/abstain interface collapses those states, whereas claim-selective certification records which claims are expressed, which evidence supports them, and which response action follows from the resulting certificate.

Shortcut controls are part of the empirical result. Intent-majority reaches 0.9013/0.8934 action accuracy, and source-conditioned action-only rows can exceed the full selector. The distinction is structural: metadata-only controls predict an action label, while the proposed interface selects claims, assigns evidence, records relation scores, and produces an auditable certificate. No-intent, intent-shuffled, and evidence-shuffled rows show sensitivity to policy inputs.

This distinction matters for interpreting the primary numbers. Action accuracy alone would make the source-conditioned majority rows a confound: they show that a large part of the weak-label action space is recoverable from source and intent metadata. The controls expose that structure and keep the empirical claim at the level of evidence-linked certification. The proposed interface is valuable because it attaches action decisions to claim-level evidence states and makes the remaining dependence on metadata measurable.

The external baselines locate the contribution. Direct answering incurs unsupported-claim risk; binary abstention reduces that risk by sacrificing utility; NLI and learned-relation claim selection are the strongest model-side comparisons. PubMedQA and source/evidence-novel cases show that remaining errors are mainly \texttt{full}/\texttt{partial}/\texttt{conflict} decisions under abstract-style evidence, rather than unsupported generation alone. The resulting picture is a measured same-source-family protocol with explicit source/evidence-novel and abstract-style transfer boundaries.

The source/evidence-novel result separates certificate behavior from action generalization. The full selector keeps UCCR at zero on the novelty slice, which means the expressed-claim certificate target remains satisfied under the weak-label metric. At the same time, action accuracy drops to 0.7805/0.7692, and the drop is concentrated in PubMedQA. The method preserves the certificate constraint on this derived boundary slice, while the intent/source policy still struggles when abstract-level \texttt{yes}/\texttt{no}/\texttt{maybe} judgments must be mapped to the drug-QA action interface. The abstention experiment completes action coverage: the source-missing counterfactual evaluates policy behavior after complete evidence removal, selector-based systems deterministically enter the abstain branch, and retrieval-only behavior cannot. The matching threshold-only result identifies the slice as coverage evidence for the fourth action. Overall, the empirical pattern supports a methodological claim. Claim-selective certification gives a structured way to expose mixed-evidence decisions, compare answer/abstain systems with native multi-action systems, and audit where source-family priors enter the policy. The results characterize a same-source-family weak-label protocol with explicit failure-boundary slices; deployment-oriented validation would require independently sampled source-disjoint data, expert adjudication, and naturally occurring abstention cases.

\section{Limitations}
\label{sec:limitations}

Several limitations follow from the evaluation design. Labels are construction-derived weak labels, and UCCR is an expressed-claim certificate metric within this protocol; the 100-item human re-audit reaches 73/100 agreement with $\kappa=0.5027$, so the paper provides weak-label protocol evidence rather than clinical-safety validation. Shortcut controls reveal source- and intent-conditioned action priors, and majority rows provide action-only comparisons without certificate production. The primary split has no exact question duplicates but does reuse sources and evidence: train-to-dev/test source-URL overlap is 0.7389/0.7555 and evidence-text overlap is 0.5541/0.5298. The derived source/evidence-novel holdout drops action accuracy to 0.7805/0.7692, mainly from PubMedQA, and the source-missing counterfactual tests only deterministic abstention under complete evidence removal. Finally, the selector is a fixed policy with intent-conditioned constants and scoped source-family priors, and template decomposition plus cue-based relation scoring still leave fine-grained multi-claim decomposition, implicit contraindications, semantic disagreement, and PubMedQA-style abstract judgments as open extensions.

\enlargethispage{2\baselineskip}

\section{Conclusion}
\label{sec:conclusion}

We study claim-selective certification as an auditable alternative to answer-or-abstain behavior for mixed-evidence medical QA. On the real-source-only protocol, the full system records UCCR$=0.0000$ on dev/test, PAU$=1.0000/0.9967$, PAU Precision$=0.9901/0.9739$, and action accuracy$=0.9204/0.8997$; shortcut controls and novelty slices define both the value and the boundary of this weak-label protocol. Rather than claiming source-disjoint clinical generalization, the paper isolates a reproducible certification interface, the shortcut structure of its weak-label action space, and the transfer boundary exposed by abstract-style evidence.

\bibliographystyle{unsrt}
\bibliography{references}

\appendix
\section{Appendix Overview}

This appendix reports the claim summary, additional diagnostics, uncertainty estimates, reproducibility commands, and asset information used to support the main paper.

\begin{algorithm}[H]
\caption{Schematic claim-selective action interface. The procedure summarizes the fixed decision flow from retrieved evidence to the final action.}
\label{alg:claim_selective_policy}
\begin{algorithmic}[1]
\STATE \textbf{Input:} question $q$, retrieved evidence $\mathcal{E}$
\STATE Emit claim skeletons $\mathcal{C}$ and question intent.
\FOR{claim $c_i\in\mathcal{C}$}
  \STATE Score support, conflict, and limitation against each $e_j\in\mathcal{E}$.
  \STATE Aggregate scores into $(S_{\mathrm{sup}},S_{\mathrm{conf}},S_{\mathrm{lim}})$.
  \STATE Map relation scores and intent to a status in \{certified, condition-limited, conflicting, omitted\}.
\ENDFOR
\STATE Build an evidence-linked certificate for expressed claims.
\IF{no claim is expressible}
  \STATE return \texttt{abstain}
\ELSIF{material conflict is selected}
  \STATE return \texttt{conflict}
\ELSIF{all required critical claims are directly certified}
  \STATE return \texttt{full}
\ELSE
  \STATE return \texttt{partial}
\ENDIF
\end{algorithmic}
\end{algorithm}

\section{Claim Summary}

\begin{table}[H]
\centering
\caption{Summary of the main claims and their evaluation level.}
\label{tab:scope_of_claims}
\scriptsize
\TableDense
\begin{tabular}{p{0.59\linewidth}p{0.33\linewidth}}
\toprule
\textbf{Claim} & \textbf{Evaluation level} \\
\midrule
Claim-selective RAG defines an action interface over certified, condition-limited, conflicting, and omitted claims & method formulation \\
Intent-aware selection improves action behavior over global thresholding & controlled weak-label dev/test comparison \\
Binary answer/abstain baselines are less expressive for mixed evidence & external-form baseline comparison \\
UCCR$=0$ is achieved by the full selector on the primary split & certificate-level metric on expressed critical claims \\
Shortcut controls expose source- and intent-conditioned action priors & metadata and perturbation diagnostics \\
\bottomrule
\end{tabular}
\end{table}

\section{Primary Split and Protocol Tables}

\begin{table}[H]
\centering
\caption{Primary real-source-only split used for the main results.}
\label{tab:appendix_clean_mainline}
\small
\TableLoose
\begin{tabular}{lrr}
\toprule
\textbf{Split} & \textbf{Size} & \textbf{Removed synthetic rows} \\
\midrule
Train & 1,470 & 84 \\
Dev eval & 314 & 18 \\
Test eval & 319 & 18 \\
\bottomrule
\end{tabular}
\end{table}

\begin{table}[H]
\centering
\caption{Gold action coverage and abstention scope. The primary dev/test protocol evaluates the non-abstain actions, and source-missing rows exercise abstention under counterfactual retrieval failure.}
\label{tab:appendix_abstain_scope}
\scriptsize
\TableDense
\begin{tabular}{lrrrrrp{0.30\linewidth}}
\toprule
\textbf{Evaluation set} & \textbf{n} & \textbf{full} & \textbf{partial} & \textbf{conflict} & \textbf{abstain} & \textbf{Role} \\
\midrule
Primary dev eval & 314 & 156 & 143 & 15 & 0 & main protocol \\
Primary test eval & 319 & 187 & 112 & 20 & 0 & main protocol test split \\
Source-missing dev stress & 314 & 0 & 0 & 0 & 314 & abstain coverage \\
Source-missing test stress & 319 & 0 & 0 & 0 & 319 & abstain coverage \\
\bottomrule
\end{tabular}
\end{table}

\section{Human Re-audit of Weak Labels}

To estimate how closely the weak-label certificate protocol tracks pharmacist adjudication, we separately re-audited a 100-item human-validation subset using official-source-first evidence review. The audit proceeded through an initial pass, a second-pass whole-set adjudication, and an item-by-item third-pass web re-audit of all remaining disagreement cases, with screenshots, HTML snapshots, and text snapshots preserved under \path{data/annotation/third_pass_web_reaudit/}. The final audited labels were not used to tune thresholds or replace the primary benchmark; they are reported only as a calibration layer for interpreting the weak-label protocol.

\begin{table}[H]
\centering
\caption{Summary of the 100-item human re-audit. Agreement is measured against the original weak labels.}
\label{tab:human_reaudit_summary}
\small
\TableStd
\begin{tabular}{lr}
\toprule
\textbf{Quantity} & \textbf{Value} \\
\midrule
Audited subset size & 100 \\
Final \texttt{full\_support} & 49 \\
Final \texttt{conditional\_support} & 49 \\
Final \texttt{conflict} & 2 \\
Weak-label agreement & 73 / 100 = 0.7300 \\
Cohen's $\kappa$ & 0.5027 \\
\bottomrule
\end{tabular}
\end{table}

\begin{table}[H]
\centering
\caption{Disagreement structure in the 100-item human re-audit. Rows count weak-label $\rightarrow$ audited-label transitions among the 27 disagreement items.}
\label{tab:human_reaudit_disagreements}
\small
\TableStd
\begin{tabular}{lr}
\toprule
\textbf{Weak $\rightarrow$ audited transition} & \textbf{Count} \\
\midrule
\texttt{conditional\_support} $\rightarrow$ \texttt{full\_support} & 11 \\
\texttt{full\_support} $\rightarrow$ \texttt{conditional\_support} & 11 \\
\texttt{conflict} $\rightarrow$ \texttt{conditional\_support} & 3 \\
\texttt{conflict} $\rightarrow$ \texttt{full\_support} & 2 \\
\bottomrule
\end{tabular}
\end{table}

Most disagreements are boundary cases between \texttt{full\_support} and \texttt{conditional\_support}, not reversals between support and conflict. The third-pass item-by-item web re-audit made no further changes to the final audited set. We therefore use this subset to calibrate scope, not to relabel the main benchmark: it supports the claim that the protocol carries nontrivial medical signal while still falling short of expert-adjudicated clinical gold data.

The source-missing counterfactual slice completes action-interface coverage. It preserves the original primary dev/test questions and claim skeletons, sets \texttt{evidence\_pool=[]} and clears \texttt{evidence\_text}, converts all gold claims to \texttt{omitted}, and sets the gold action to \texttt{abstain}. Table~\ref{tab:appendix_abstain_stress_results} reports the corresponding action behavior. Because every stress item has empty evidence, these rows characterize deterministic abstention under retrieval failure; semantic evidence insufficiency, natural abstention frequency, and expert-labeled correctness require separate evaluation data.

\begin{table}[H]
\centering
\caption{Source-missing abstention results. Selector rows abstain under empty evidence, whereas retrieval/relation-only rows express claims without evidence and fail the action check.}
\label{tab:appendix_abstain_stress_results}
\small
\TableStd
\begin{tabular}{llccc}
\toprule
\textbf{Split} & \textbf{Configuration} & \textbf{UCCR} & \textbf{Action Acc} & \textbf{Gold abstain} \\
\midrule
\multirow{4}{*}{Dev ($n=314$)}
& Retrieval only & 1.0000 & 0.0000 & 314 \\
& Relation only & 1.0000 & 0.0000 & 314 \\
& Threshold-only selector & 0.0000 & 1.0000 & 314 \\
& Full risk-calibrated & \textbf{0.0000} & \textbf{1.0000} & 314 \\
\midrule
\multirow{4}{*}{Test ($n=319$)}
& Retrieval only & 1.0000 & 0.0000 & 319 \\
& Relation only & 1.0000 & 0.0000 & 319 \\
& Threshold-only selector & 0.0000 & 1.0000 & 319 \\
& Full risk-calibrated & \textbf{0.0000} & \textbf{1.0000} & 319 \\
\bottomrule
\end{tabular}
\end{table}

\begin{table}[H]
\centering
\caption{Information available to weak-label construction and model-side policies.}
\label{tab:appendix_label_policy_info}
\scriptsize
\TableDense
\begin{tabular}{lccc}
\toprule
\textbf{Component} & \textbf{Weak labels} & \textbf{Selector} & \textbf{Note} \\
\midrule
Question text & yes & yes & shared task input \\
Retrieved evidence & yes & yes & shared RAG setting \\
Weak label / gold action & yes & no & evaluation target only \\
Question intent & claim skeleton & yes & tested by intent controls \\
Source type / root claim type & label construction & scoped priors & tested by source-conditioned controls \\
Relation scores & no & yes & tested by no-intent and shuffled controls \\
Dev labels & tuning & tuning controls & transferred to test \\
Test labels & evaluation & no & reporting only \\
\bottomrule
\end{tabular}
\end{table}

\begin{table}[H]
\centering
\caption{Shortcut controls on the primary split. Majority rows are action-only controls fit from training-set weak labels, and certificate metrics apply only to rows with evidence-linked claim outputs.}
\label{tab:appendix_shortcut_controls}
\small
\TableStd
\begin{tabular}{lcc}
\toprule
\textbf{Control} & \textbf{Dev Action Acc} & \textbf{Test Action Acc} \\
\midrule
No-intent threshold selector & 0.5223 & 0.5517 \\
Train intent-majority only & 0.9013 & 0.8934 \\
Train source+intent majority only & 0.9299 & 0.9310 \\
Train source+claim-type majority only & 0.9331 & 0.9279 \\
Evidence-shuffled full selector & 0.8344 & 0.8088 \\
Intent-shuffled full selector & 0.4936 & 0.4671 \\
Full intent-aware selector & 0.9204 & 0.8997 \\
\bottomrule
\end{tabular}
\end{table}

\begin{table}[H]
\centering
\caption{Baseline package on the primary split. Binary-form baselines use the mapping \texttt{answer}$\rightarrow$\texttt{full} and \texttt{abstain}$\rightarrow$\texttt{abstain} for action accuracy, whereas claim-selective rows retain the native \{full, partial, conflict, abstain\} action space.}
\label{tab:appendix_baseline_package}
\small
\TableLoose
\begin{tabular}{llcccc}
\toprule
\textbf{Split} & \textbf{Method} & \textbf{UCCR} & \textbf{PAU} & \textbf{F1} & \textbf{Action Acc} \\
\midrule
\multirow{8}{*}{Dev ($n=314$)}
& vanilla answer & 0.2070 & 0.7860 & 0.9755 & 0.4968 \\
& binary answer / abstain & 0.0000 & 0.7793 & 0.8535 & 0.4076 \\
& verifier only & 0.0000 & 0.7659 & 0.8450 & 0.3949 \\
& citation only & 0.2070 & 0.7860 & 0.9755 & 0.4968 \\
& speech-act-guided proxy & 0.0000 & 0.9732 & 0.9652 & 0.4777 \\
& claim-selective + NLI relation & 0.0000 & 0.8595 & 0.9179 & 0.7484 \\
& claim-selective + learned relation & 0.0000 & 0.9766 & 0.9815 & 0.8758 \\
& claim-selective full & \textbf{0.0000} & \textbf{1.0000} & \textbf{0.9950} & \textbf{0.9204} \\
\midrule
\multirow{8}{*}{Test ($n=319$)}
& vanilla answer & 0.2194 & 0.7692 & 0.9676 & 0.5862 \\
& binary answer / abstain & 0.0000 & 0.7659 & 0.8373 & 0.4671 \\
& verifier only & 0.0000 & 0.7525 & 0.8287 & 0.4545 \\
& citation only & 0.2194 & 0.7692 & 0.9676 & 0.5862 \\
& speech-act-guided proxy & 0.0000 & 0.9599 & 0.9551 & 0.5611 \\
& claim-selective + NLI relation & 0.0000 & 0.8094 & 0.8768 & 0.6614 \\
& claim-selective + learned relation & 0.0000 & 0.9900 & 0.9801 & 0.8652 \\
& claim-selective full & \textbf{0.0000} & \textbf{0.9967} & \textbf{0.9851} & \textbf{0.8997} \\
\bottomrule
\end{tabular}
\end{table}

\section{Extended Diagnostics}

This section expands the main source- and claim-type analyses for the full system. The tables and figures below report diagnostic slices under the same weak-label protocol as the main results.

\begin{table}[H]
\centering
\caption{Source-level diagnostics for the full system on the primary split.}
\label{tab:appendix_source_slices}
\small
\TableLoose
\begin{tabular}{lrrrrrr}
\toprule
\multirow{2}{*}{\textbf{Source}} & \multicolumn{3}{c}{\textbf{Dev}} & \multicolumn{3}{c}{\textbf{Test}} \\
\cmidrule(lr){2-4} \cmidrule(lr){5-7}
& \textbf{n} & \textbf{PAU} & \textbf{Action Acc} & \textbf{n} & \textbf{PAU} & \textbf{Action Acc} \\
\midrule
OpenFDA & 208 & 1.0000 & 0.9663 & 209 & 0.9949 & 0.9330 \\
FDA FAERS & 40 & 1.0000 & 1.0000 & 41 & 1.0000 & 1.0000 \\
PubMed Literature & 37 & 1.0000 & 0.8919 & 38 & 1.0000 & 0.9211 \\
PubMedQA & 29 & 1.0000 & 0.5172 & 31 & 1.0000 & 0.5161 \\
\bottomrule
\end{tabular}
\end{table}

\begin{figure}[H]
\centering
\includegraphics[width=0.92\linewidth]{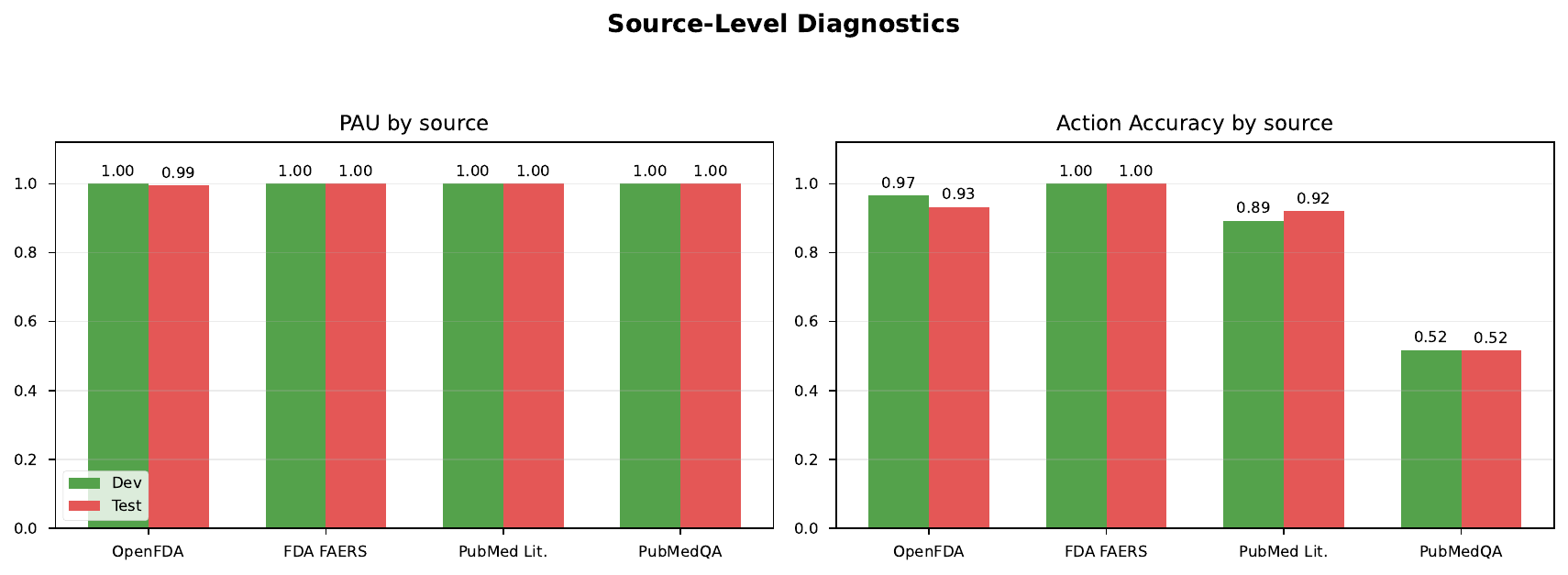}
\caption{Extended source-level diagnostics for the full system. OpenFDA has the highest action accuracy, whereas PubMedQA has the lowest.}
\label{fig:appendix_source_diagnostics}
\end{figure}

\begin{table}[H]
\centering
\caption{Hard claim-type diagnostics for the full system. Slices with $n<20$ are reported descriptively.}
\label{tab:appendix_weak_slices}
\small
\TableStd
\begin{tabular}{lcccccc}
\toprule
\multirow{2}{*}{\textbf{Claim type}} & \multicolumn{3}{c}{\textbf{Dev}} & \multicolumn{3}{c}{\textbf{Test}} \\
\cmidrule(lr){2-4} \cmidrule(lr){5-7}
& \textbf{n} & \textbf{PAU} & \textbf{Action Acc} & \textbf{n} & \textbf{PAU} & \textbf{Action Acc} \\
\midrule
Indication & 29 & 1.0000 & 0.5172 & 31 & 1.0000 & 0.5161 \\
Pregnancy & 11 & 1.0000 & 0.7273 & 14 & 0.9231 & 0.5714 \\
Lactation & 5 & 1.0000 & 0.8000 & 7 & 1.0000 & 0.7143 \\
Interaction & 19 & 1.0000 & 0.8421 & 17 & 1.0000 & 0.7647 \\
Dosage adjustment & 11 & 1.0000 & 0.9091 & 11 & 1.0000 & 0.8182 \\
Dosage & 6 & 1.0000 & 1.0000 & 12 & 1.0000 & 0.9167 \\
Special population & 8 & 1.0000 & 0.7500 & 9 & 1.0000 & 1.0000 \\
\bottomrule
\end{tabular}
\end{table}

\begin{figure}[H]
\centering
\includegraphics[width=0.82\textwidth]{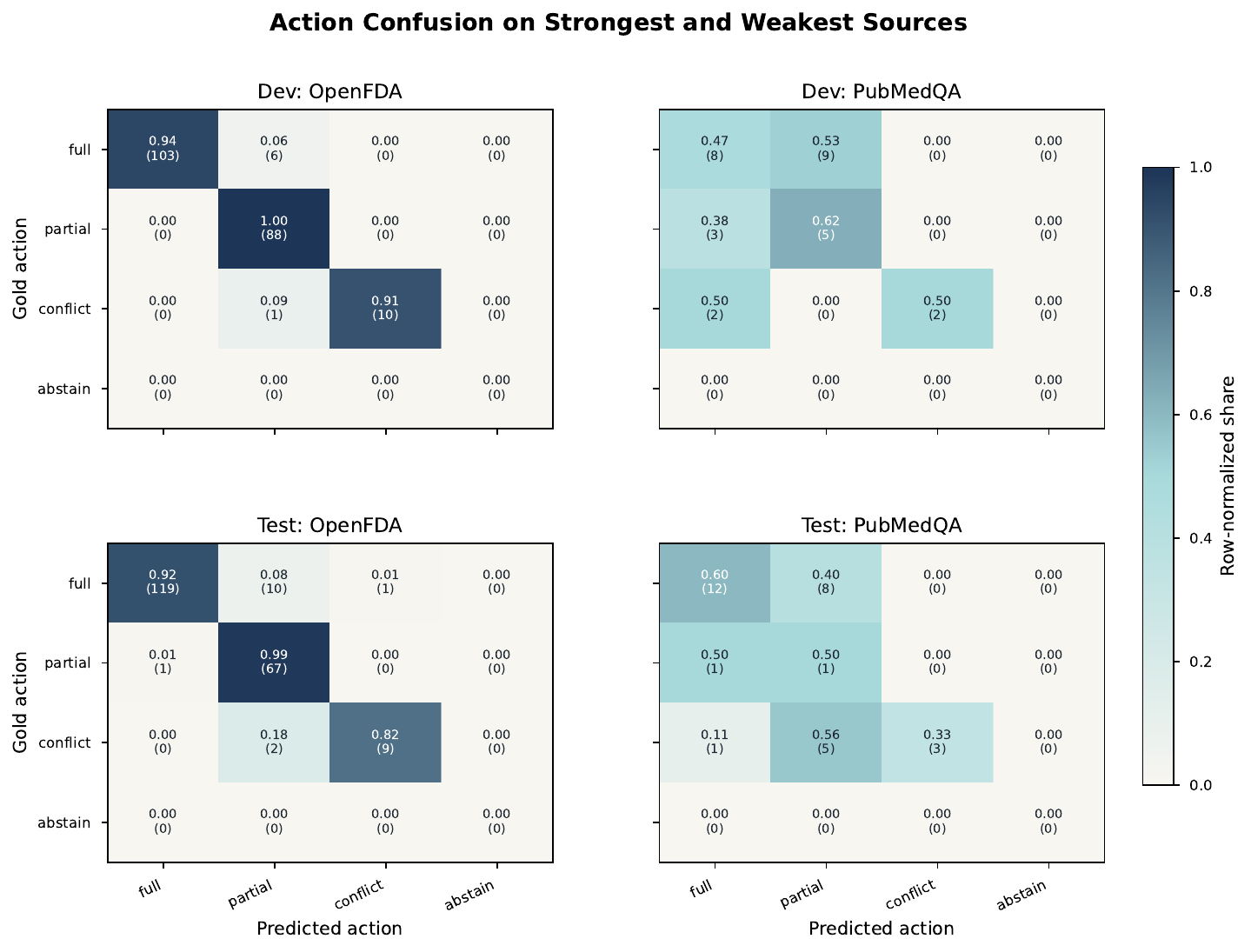}
\caption{Action confusion on the strongest and weakest sources. PubMedQA errors are dominated by \texttt{full}/\texttt{partial} boundary mistakes rather than unsupported generation.}
\label{fig:appendix_action_confusions}
\end{figure}

\section{Source-Overlap Analysis}
\label{sec:source_overlap_audit}

The primary split removes exact question duplicates and retains source-family overlap. We measure exact overlap over question text, source URL, and evidence text. Table~\ref{tab:source_overlap_audit} reports the main analysis. Train-to-dev/test question overlap is zero, while source-URL overlap is substantial: 232/314 dev examples and 241/319 test examples share a source URL with training. Evidence-text overlap is also nontrivial at 174/314 and 169/319. The source pattern is uneven: OpenFDA examples have 100\% train source-URL overlap in both dev and test, FDA FAERS has partial overlap, and PubMed Literature/PubMedQA have no exact train source-URL overlap.

\begin{table}[H]
\centering
\caption{Source-overlap analysis for the primary split. Novelty-slice metrics use the unchanged full selector, and the source/evidence-novel holdout requires both source URL and evidence text to be absent from train.}
\label{tab:source_overlap_audit}
\small
\TableStd
\begin{tabular}{llrrrr}
\toprule
\textbf{Diagnostic} & \textbf{Split / key} & \textbf{n} & \textbf{Overlap} & \textbf{Rate} & \textbf{Action Acc} \\
\midrule
Exact overlap & train$\rightarrow$dev question & 314 & 0 & 0.0000 & -- \\
Exact overlap & train$\rightarrow$dev source URL & 314 & 232 & 0.7389 & -- \\
Exact overlap & train$\rightarrow$dev evidence text & 314 & 174 & 0.5541 & -- \\
Exact overlap & train$\rightarrow$test question & 319 & 0 & 0.0000 & -- \\
Exact overlap & train$\rightarrow$test source URL & 319 & 241 & 0.7555 & -- \\
Exact overlap & train$\rightarrow$test evidence text & 319 & 169 & 0.5298 & -- \\
\midrule
Full selector & dev all & 314 & -- & -- & 0.9204 \\
Full selector & dev source/evidence-novel & 82 & -- & -- & 0.7805 \\
Full selector & test all & 319 & -- & -- & 0.8997 \\
Full selector & test source/evidence-novel & 78 & -- & -- & 0.7692 \\
\bottomrule
\end{tabular}
\end{table}

The source/evidence-novel holdout is smaller and distribution-shifted. Its lower action accuracy characterizes the primary split as a controlled same-source-family evaluation with a separate source/evidence-novel transfer slice. On this boundary slice, the full selector records UCCR$=0.0000$ under the weak-label certificate metric, with PAU/F1 of 1.0000/0.9873 on dev and 1.0000/0.9583 on test. Tables~\ref{tab:source_novel_holdout} and~\ref{tab:source_novel_diagnostics} provide the reproducible slice definition and the corresponding comparison rows.

\begin{table}[H]
\centering
\caption{Source/evidence-novel holdout construction. Rows are selected from the primary dev/test evaluation files when both normalized \texttt{source\_url} and \texttt{evidence\_text} are absent from train.}
\label{tab:source_novel_holdout}
\small
\TableStd
\begin{tabular}{llrrrr}
\toprule
\textbf{Split} & \textbf{Distribution} & \textbf{n} & \textbf{full} & \textbf{partial} & \textbf{conflict} \\
\midrule
Dev & all source/evidence-novel & 82 & 32 & 46 & 4 \\
Dev & PubMed Literature / PubMedQA / FDA FAERS & 82 & \multicolumn{3}{c}{37 / 29 / 16} \\
Test & all source/evidence-novel & 78 & 31 & 38 & 9 \\
Test & PubMed Literature / PubMedQA / FDA FAERS & 78 & \multicolumn{3}{c}{38 / 31 / 9} \\
\bottomrule
\end{tabular}
\end{table}

\begin{table}[H]
\centering
\caption{Source/evidence-novel holdout results. Selector and baseline rows use fixed primary-split settings with no threshold retuning, and majority rows are action-only controls that do not produce certificates.}
\label{tab:source_novel_diagnostics}
\scriptsize
\TableDense
\begin{tabular}{llccccc}
\toprule
\textbf{Split} & \textbf{Configuration} & \textbf{UCCR} & \textbf{PAU} & \textbf{F1} & \textbf{Action Acc} & \textbf{Role} \\
\midrule
\multirow{7}{*}{Dev ($n=82$)}
& Threshold-only selector & 0.0000 & 1.0000 & 0.9873 & 0.5854 & fixed selector \\
& Full risk-calibrated & \textbf{0.0000} & \textbf{1.0000} & \textbf{0.9873} & 0.7805 & certificate \\
& Learned relation + selector & 0.0000 & 0.9103 & 0.9342 & 0.6341 & module swap \\
& NLI relation + selector & 0.0000 & 0.7949 & 0.8671 & 0.5488 & NLI baseline \\
& Source+intent majority & -- & -- & -- & 0.8293 & action-only \\
& Source+claim-type majority & -- & -- & -- & 0.8415 & action-only \\
& Evidence-shuffled full & 0.0000 & 1.0000 & 0.9750 & 0.7073 & perturbation \\
\midrule
\multirow{7}{*}{Test ($n=78$)}
& Threshold-only selector & 0.0000 & 0.9420 & 0.9420 & 0.5256 & fixed selector \\
& Full risk-calibrated & \textbf{0.0000} & \textbf{1.0000} & \textbf{0.9583} & 0.7692 & certificate \\
& Learned relation + selector & 0.0000 & 0.9565 & 0.9296 & 0.5769 & module swap \\
& NLI relation + selector & 0.0000 & 0.8551 & 0.8613 & 0.4872 & NLI baseline \\
& Source+intent majority & -- & -- & -- & 0.8333 & action-only \\
& Source+claim-type majority & -- & -- & -- & 0.8205 & action-only \\
& Evidence-shuffled full & 0.0000 & 0.9710 & 0.9241 & 0.6282 & perturbation \\
\bottomrule
\end{tabular}
\end{table}

\begin{table}[H]
\centering
\caption{Source/evidence-novel full-selector accuracy by source. The non-PubMedQA aggregate shows that most of the action drop is localized to the abstract-style PubMedQA transfer slice.}
\label{tab:source_novel_by_source}
\small
\TableLoose
\begin{tabular}{lrrrr}
\toprule
\multirow{2}{*}{\textbf{Source group}} & \multicolumn{2}{c}{\textbf{Dev}} & \multicolumn{2}{c}{\textbf{Test}} \\
\cmidrule(lr){2-3} \cmidrule(lr){4-5}
& \textbf{n} & \textbf{Action Acc} & \textbf{n} & \textbf{Action Acc} \\
\midrule
FDA FAERS & 16 & 1.0000 & 9 & 1.0000 \\
PubMed Literature & 37 & 0.8919 & 38 & 0.9211 \\
PubMedQA & 29 & 0.5172 & 31 & 0.5161 \\
Non-PubMedQA aggregate & 53 & 0.9245 & 47 & 0.9362 \\
\bottomrule
\end{tabular}
\end{table}

On the source/evidence-novel holdout, the full selector makes 18 action errors on dev and 18 on test. The source-specific pattern is concentrated: FDA FAERS remains at 1.0000 action accuracy on both slices, PubMed Literature remains high (0.8919/0.9211), and PubMedQA remains low (0.5172/0.5161). Excluding PubMedQA, source/evidence-novel action accuracy is 0.9245/0.9362. This supports interpreting the holdout as an abstract-style and source-shift boundary rather than a broad failure of the certificate metric.

\section{Statistical Uncertainty}

We computed nonparametric bootstrap intervals on the primary real-source-only dev and test splits. Table~\ref{tab:appendix_bootstrap} reports point estimates and 95\% intervals for threshold-only selection and the full intent-aware selector.

The threshold-only comparison uses the \textit{dev-selected} operating point described in Section~\ref{sec:experiments}. A grid search over \texttt{support}, \texttt{conflict}, and \texttt{condition\_limited} thresholds selects the candidate with UCCR$=0$ and then maximizes PAU, action accuracy, and F1 in that order. The selected setting is \texttt{support}=0.35, \texttt{conflict}=0.55, and \texttt{condition\_limited}=0.30, transferred unchanged to test. The speech-act-guided answer/abstain proxy is tuned separately on dev by searching a small set of global \texttt{answer\_support} gates while keeping its profile-specific retrieval logic fixed; this selects \texttt{answer\_support}=0.34.

\begin{table}[H]
\centering
\caption{Bootstrap 95\% intervals for the main selector comparison. Intervals are estimated from 1,000 sample-level bootstrap resamples.}
\label{tab:appendix_bootstrap}
\small
\TableLoose
\resizebox{\textwidth}{!}{%
\begin{tabular}{llcccc}
\toprule
\textbf{Split} & \textbf{Method} & \textbf{UCCR} & \textbf{PAU} & \textbf{F1} & \textbf{Action Acc} \\
\midrule
Dev & threshold-only & 0.0000 [0.0000, 0.0000] & 0.9933 [0.9832, 1.0000] & 0.9754 [0.9633, 0.9870] & 0.5223 [0.4682, 0.5732] \\
Dev & full selector & 0.0000 [0.0000, 0.0000] & 1.0000 [1.0000, 1.0000] & 0.9950 [0.9884, 1.0000] & 0.9204 [0.8854, 0.9490] \\
Test & threshold-only & 0.0000 [0.0000, 0.0000] & 0.9732 [0.9529, 0.9900] & 0.9620 [0.9456, 0.9756] & 0.5517 [0.4953, 0.6050] \\
Test & full selector & 0.0000 [0.0000, 0.0000] & 0.9967 [0.9870, 1.0000] & 0.9851 [0.9749, 0.9935] & 0.8997 [0.8652, 0.9310] \\
\bottomrule
\end{tabular}
}
\end{table}

The paired bootstrap deltas localize the gain. On dev, the full selector improves PAU by 0.0067 with a 95\% interval of [0.0000, 0.0168], F1 by 0.0196 [0.0097, 0.0299], and action accuracy by 0.3981 [0.3408, 0.4522]. On test, the corresponding deltas are 0.0234 [0.0068, 0.0435] for PAU, 0.0231 [0.0099, 0.0381] for F1, and 0.3480 [0.2915, 0.4075] for action accuracy.

For UCCR, bootstrap intervals degenerate at zero because no unsupported expressed critical claim is observed under the reported weak-label certificate metric on the primary split. Interpreting the dev/test expressed critical claims as Bernoulli trials gives 0 observed events out of 302 expressed critical claims on dev and 0 out of 306 on test. The corresponding upper ends of the two-sided 95\% Wilson score intervals are 0.0126 and 0.0124.

\section{PAU Precision}

PAU is a recall-style utility metric: it measures how many gold usable claims are retained. To make the over-expression boundary explicit, we also compute a precision counterpart,
$|\mathrm{gold\ usable}\cap \mathrm{pred\ usable}|/|\mathrm{pred\ usable}|$.
This precision counterpart distinguishes retained useful claims from over-expression.

\begin{table}[H]
\centering
\caption{PAU precision on the primary split. Higher values indicate fewer extra usable predictions beyond the weak-label usable set.}
\label{tab:pau_precision_audit}
\small
\TableStd
\begin{tabular}{lcc}
\toprule
\textbf{Method} & \textbf{Dev} & \textbf{Test} \\
\midrule
Retrieval only & 0.9522 & 0.9373 \\
Threshold-only selector & 0.9581 & 0.9510 \\
Full selector & 0.9901 & 0.9739 \\
\bottomrule
\end{tabular}
\end{table}

Retrieval-only keeps PAU at 1.0000 because it expresses all gold usable claims, but its PAU precision is lower because it also expresses claims outside the weak-label usable set. The full selector has the strongest PAU precision among these rows, indicating that its high PAU is not obtained by broad over-expression.

\section{Selector Threshold Sensitivity}

We run one-at-a-time perturbations around the reported selector operating points. For the tuned threshold-only selector, we perturb \texttt{support}, \texttt{conflict}, \texttt{condition\_limited}, and \texttt{limitation} by $\pm 0.05$ from the dev-selected point. For the full selector, we apply the same perturbation to its global fallback thresholds while leaving the intent-conditioned policy branches fixed.

\begin{table}[H]
\centering
\caption{Selector threshold sensitivity. All rows have UCCR$=0.0000$ under the reported weak-label certificate metric, and the full selector is unchanged because the high-impact decisions are governed by intent-conditioned policy branches.}
\label{tab:appendix_selector_sensitivity}
\small
\TableLoose
\begin{tabular}{llccc}
\toprule
\textbf{Split} & \textbf{Selector variant} & \textbf{PAU} & \textbf{F1} & \textbf{Action Acc} \\
\midrule
Dev & threshold-only base & 0.9933 & 0.9754 & 0.5223 \\
Dev & threshold-only, \texttt{condition\_limited}+0.05 & 0.8161 & 0.8777 & 0.4076 \\
Dev & full selector base & 1.0000 & 0.9950 & 0.9204 \\
Dev & full selector, any one global threshold $\pm$0.05 & 1.0000 & 0.9950 & 0.9204 \\
\midrule
Test & threshold-only base & 0.9732 & 0.9620 & 0.5517 \\
Test & threshold-only, \texttt{condition\_limited}+0.05 & 0.8094 & 0.8721 & 0.4671 \\
Test & full selector base & 0.9967 & 0.9851 & 0.8997 \\
Test & full selector, any one global threshold $\pm$0.05 & 0.9967 & 0.9851 & 0.8997 \\
\bottomrule
\end{tabular}
\end{table}

The sensitivity analysis matches the main result table. Threshold-only behavior can keep UCCR at zero, but its utility and action behavior are brittle to a modest increase in the condition-limited gate. The full selector is dominated by intent-conditioned branches, so these global fallback-threshold perturbations leave its dev/test metrics unchanged. The result localizes the reported decisions to the intent-conditioned policy branches rather than to the global fallback gates.

\section{Selector Policy Specification}
\label{sec:selector_policy_audit}

The full selector is a fixed policy specification with intent-conditioned branches and global fallback thresholds. The repository includes a policy-constant audit at \path{scripts/audit_selector_policy_constants.py} for \path{src/selection/selector.py}. The audit covers 445 lines across the fallback, intent-aware, and final classification functions and finds 141 numeric constants in those functions. These constants define the implemented policy for the current relation-score distribution; shortcut controls, source slices, and threshold perturbations characterize the sensitivity of that policy.

We additionally run a branch-family sensitivity audit at \path{scripts/analyze_selector_branch_sensitivity.py}. This audit perturbs the support, conflict, or limitation score entering one intent branch family at a time by $\pm 0.05$ and $\pm 0.10$ to characterize robustness of the fixed policy. The largest overall action-accuracy drops are $-0.0382$ on dev and $-0.0408$ on test, both under a $-0.10$ support perturbation to the indication branch. Slice-level changes are larger in small or boundary-heavy families, including interaction on dev and dosage-adjustment and research-question slices on test. In the largest-change rows, UCCR remains $0.0000$ under the certificate metric. This audit localizes sensitivity to specific intent-conditioned branches rather than to the global fallback gates.

\section{Selective-Prediction View}

We also report a risk--coverage view at the critical-claim level. Each critical claim is treated as a selectable item. Coverage is selected critical claims divided by total gold critical claims, and risk is selected critical claims not marked supportable by the gold weak labels divided by selected critical claims. This view is supplementary because it ignores conflict disclosure and certificate semantics.

For the threshold-only selector, we sweep the support threshold and tie the condition-limited threshold to \texttt{support - 0.10} while keeping the conflict thresholds fixed. We also plot the dev-tuned threshold-only operating point (\texttt{support}=0.35, \texttt{conflict}=0.55, \texttt{condition\_limited}=0.30). The full selector, learned claim-selective baseline, and NLI baseline are reported as fixed operating points. On dev, the tuned threshold-only point reaches coverage 0.9873 and risk 0.0419, whereas the full selector reaches coverage 0.9618 and risk 0.0099. On test, the corresponding comparison is 0.9592 / 0.0490 for the tuned threshold-only point versus 0.9592 / 0.0261 for the full selector. The learned claim-selective baseline remains close but is slightly lower-coverage and higher-risk than the full selector on both splits (0.9427/0.9561 coverage and 0.0135/0.0295 risk).

\begin{figure}[H]
\centering
\includegraphics[width=0.92\linewidth]{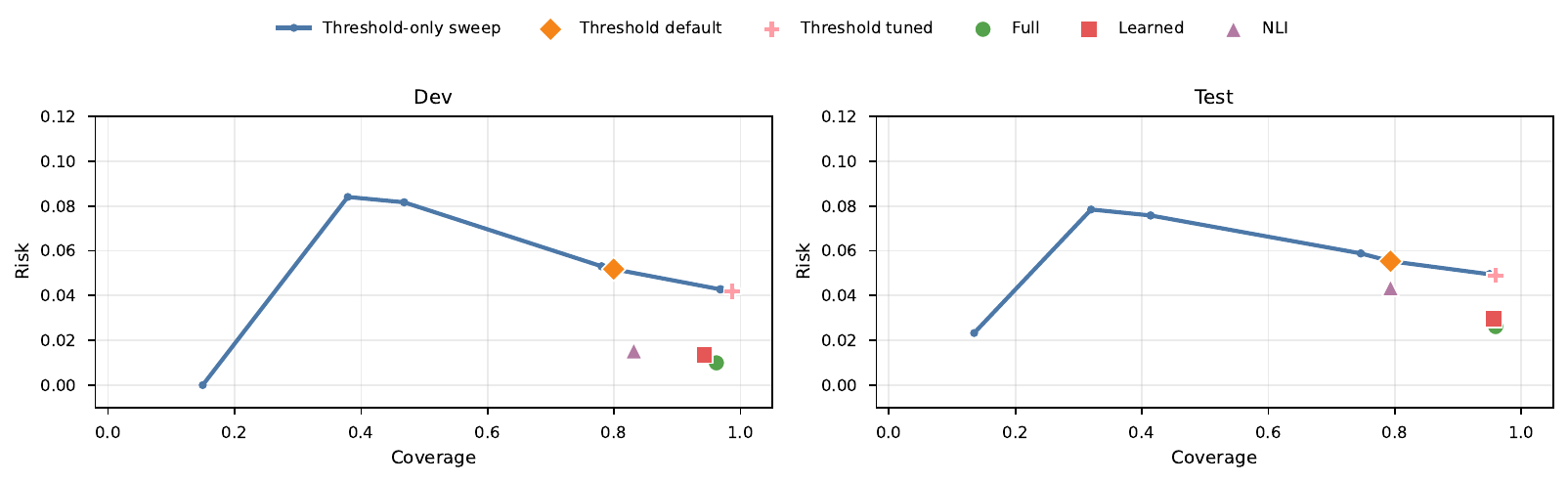}
\caption{Selective-prediction view on the primary split. The threshold-only selector traces a tunable risk--coverage family, and the full selector keeps similar coverage but lower risk than the dev-tuned threshold-only operating point on both splits.}
\label{fig:appendix_risk_coverage}
\end{figure}

\section{Compute Resources}

Experiments were reproduced on a Windows workstation with an Intel Core Ultra 9 185H CPU (16 physical cores / 22 logical processors), approximately 33.95 GB of physical memory, Intel Arc integrated graphics, and an NVIDIA GeForce RTX 4060 Laptop GPU. The main runtime costs come from relation-model loading and repeated baseline sweeps rather than large-scale training.

Representative wall-clock times are:
\begin{itemize}
    \item dev ablation rerun: 110.93 seconds
    \item baseline package rerun on the primary real-source-only split: 321.94 seconds
\end{itemize}

\section{Existing Assets and Terms}

The reported evaluation chain depends on the following external assets.
\begin{itemize}
    \item \textbf{openFDA / FAERS APIs.} The official openFDA terms state that, unless otherwise noted, content and data are generally unrestricted and made available under a CC0 1.0 dedication, while also warning that some third-party content may be separately marked and that API use is subject to service limits and terms.\footnote{\url{https://open.fda.gov/terms}}
    \item \textbf{DailyMed.} DailyMed is provided by the U.S. National Library of Medicine as the official public source of FDA label information, but the site also notes that NLM does not review SPL content before publication and that the ``in use'' labeling may differ from the most recent FDA-approved labeling.\footnote{\url{https://dailymed.nlm.nih.gov/dailymed/about-dailymed.cfm}}
    \item \textbf{PubMed / NLM literature content.} NLM's own copyright guidance states that some NLM data are U.S. government works while abstracts and other contributed materials may still be protected by copyright, leaving downstream users responsible for respecting those restrictions when redistributing content.\footnote{\url{https://www.nlm.nih.gov/databases/download.html}}
    \item \textbf{PubMedQA.} The official PubMedQA repository distributes the benchmark under the MIT license and specifies the dataset download and evaluation process.\footnote{\url{https://github.com/pubmedqa/pubmedqa}}
    \item \textbf{NLI baseline model.} The \texttt{facebook/bart-large-mnli} model card lists the model under the MIT license.\footnote{\url{https://huggingface.co/facebook/bart-large-mnli}}
\end{itemize}

The evaluation distinguishes between public-domain or permissive API access, NLM-hosted content that may still contain copyrighted abstracts, and benchmark/model assets distributed under repository-specific licenses.

\section{Reproducibility Note}

The experiments use a single real-source-only split family:
\path{data/splits/primary_real_source/}, which contains
\path{train.jsonl}, \path{dev_eval.jsonl}, and \path{test_eval.jsonl}. The
main reported results use only the \path{dev_eval.jsonl} and \path{test_eval.jsonl} files
from this real-source-only split.

\paragraph{Main ablation reruns.}
The main ablation table can be regenerated with:
\begin{verbatim}
python scripts/run_ablation_study.py \
  data/splits/primary_real_source/dev_eval.jsonl \
  outputs/regression_checks/<dev_run>

python scripts/run_ablation_study.py \
  data/splits/primary_real_source/test_eval.jsonl \
  outputs/regression_checks/<test_run>
\end{verbatim}

\paragraph{Baseline package reruns.}
The external-form baseline package can be regenerated with:
\begin{verbatim}
python scripts/run_current_baseline_package.py \
  --split-dir data/splits/primary_real_source \
  --output-dir outputs/baselines/current_clean_mainline
\end{verbatim}
The speech-act-guided proxy tuning run can be regenerated with:
\begin{verbatim}
python scripts/tune_pragaura_proxy_baseline.py
\end{verbatim}

\paragraph{Slice diagnostics and consistency checks.}
Source-level and claim-type diagnostics are regenerated from the same split family,
and the repository includes diagnostic scripts for the data and
gapfill state:
\begin{verbatim}
python scripts/analyze_eval_slices.py --input <split_jsonl> ...
python scripts/analyze_shortcut_controls.py
python scripts/audit_source_overlap.py
python scripts/audit_selector_policy_constants.py
python scripts/audit_data_experiment_state.py
python scripts/audit_gapfill_state.py
\end{verbatim}

\paragraph{Reproducibility package.}
The commands above rerun the reported implementation inside the project repository and define the split files, frozen outputs, and regeneration commands used by the experiments. The anonymous supplemental package is a curated whitelist of the scripts, split files, stress slices, canonical outputs, and diagnostics needed to inspect the paper's claims; non-primary workflow holdouts, unfinished manual-audit preparation folders, and historical diagnostic snapshots are intentionally excluded. The package also includes a Croissant metadata file, \path{artifact/croissant_metadata.json}, with core dataset fields and Responsible AI notes for data collection, weak-label annotation, known biases, limitations, and intended research use.

\section{Synthetic Stress Slice}

The synthetic stress slice is reported separately from the main result table. On the 20-example synthetic stress set, the full system obtains UCCR$=0.0000$, PAU$=0.0000$, F1$=0.0000$, and action accuracy$=0.6500$. The zero PAU and F1 values indicate a label-interface mismatch, so the slice functions as a stress test while the real-source-only chain remains the primary result source.

\end{document}